\documentclass[conference]{IEEEtran}
\IEEEoverridecommandlockouts
\usepackage{cite}
\usepackage{amsmath,amssymb,amsfonts}
\usepackage{algorithmic}
\usepackage{graphicx}
\usepackage{textcomp}
\usepackage{xcolor}
\def\BibTeX{{\rm B\kern-.05em{\sc i\kern-.025em b}\kern-.08em
    T\kern-.1667em\lower.7ex\hbox{E}\kern-.125emX}}

\usepackage{bm,amssymb,amsmath,xspace,enumitem,multirow,booktabs,algorithm,algorithmic}

\newlist{tasklist}{enumerate}{1}
\setlist[tasklist]{%
    label=\arabic*.,
    leftmargin=2.2em,
    labelsep=0.7em,
    itemsep=6pt,      
    topsep=3pt,        
    parsep=0pt,
    partopsep=0pt
}

\newcommand*{\eg}{{\em e.g.}\@\xspace}

\newcommand{\Rmnum}[1]{\uppercase\expandafter{\romannumeral #1}}

\begin{document}

\title{Can Vision-Language-Action Models Learn from Real-World Data
Continually without Forgetting?
}

\author{
Jiarun Zhu\textsuperscript{1,3,*}, Yijun Hong\textsuperscript{1,5,*}, Xiaoquan Sun\textsuperscript{2,4,*}, Zetian Xu\textsuperscript{1,2}, Qijun He\textsuperscript{1}, Haijier Chen\textsuperscript{1,2}, Zhiyong Wang\textsuperscript{6},\\Mingqi Yuan\textsuperscript{1,2,$\dagger$}, Wenjun Zeng\textsuperscript{3,$\dagger$}, Jiayu Chen\textsuperscript{1,2}

\\
\textsuperscript{1}HKU\quad\textsuperscript{2}INFIFORCE\quad\textsuperscript{3}EIT, Ningbo\quad\textsuperscript{4}HUST\quad\textsuperscript{5}SUSTech\quad\textsuperscript{6}HITSZ
\thanks{$*$ These authors contributed equally.}
\thanks{$\dagger$ Corresponding authors: Mingqi Yuan ({\tt\small my017@hku.hk}) and Wenjun Zeng ({\tt\small wzeng-vp@eitech.edu.cn}).}
}

\maketitle

\begin{abstract}
Vision-Language-Action (VLA) models provide a promising foundation for general-purpose robotics, yet their real-world deployment demands the ability to continually acquire new skills without forgetting prior ones. While recent studies have explored continual learning for VLA models in simulated settings, the challenge remains largely unexamined under realistic physical conditions. To bridge this gap, we construct a real-world continual learning benchmark comprising ten diverse sequential manipulation tasks across both single-arm and bimanual configurations. Through extensive experiments on this benchmark, we find that naive sequential fine-tuning leads to severe catastrophic forgetting, whereas a well-configured experience replay (ER) approach can effectively mitigate forgetting and outperform joint multi-task training under equivalent computational budgets. Notably, by synthesizing our empirical findings, we successfully achieve stable continual learning across the full 10-task heterogeneous stream, retaining previously acquired capabilities while adapting to diverse new skills in real-world deployment. This work presents an empirical study grounded in real-world continual VLA learning and offers actionable insights for deploying robust, long-lived robotic policies. 
\end{abstract}


\section{Introduction}

Vision-Language-Action (VLA) models have emerged as a promising paradigm for general-purpose robotics by unifying visual perception, language understanding, and action generation within a single policy \cite{ma2024survey,kim2025openvla}. Pretrained on large-scale robotic datasets, these models exhibit impressive cross-task generalization and transfer capabilities for embodied control \cite{black2025pi}. However, practical deployment demands capabilities beyond static, single-stage performance, where robots must continually acquire new skills from non-stationary data streams without degrading previously learned abilities. This requirement is closely related to catastrophic forgetting, a significant challenge in continual learning, in which sequential model updates unintentionally erase prior knowledge \cite{de2021continual,wang2024comprehensive}. While naively updating VLA models on new task streams causes severe skill degradation, periodically retraining them from scratch on the entire accumulated dataset is computationally prohibitive. Consequently, developing effective mechanisms to mitigate catastrophic forgetting is essential for scalable and lifelong robot learning.

\begin{figure}[t!]
    \centering
    \includegraphics[width=\linewidth]{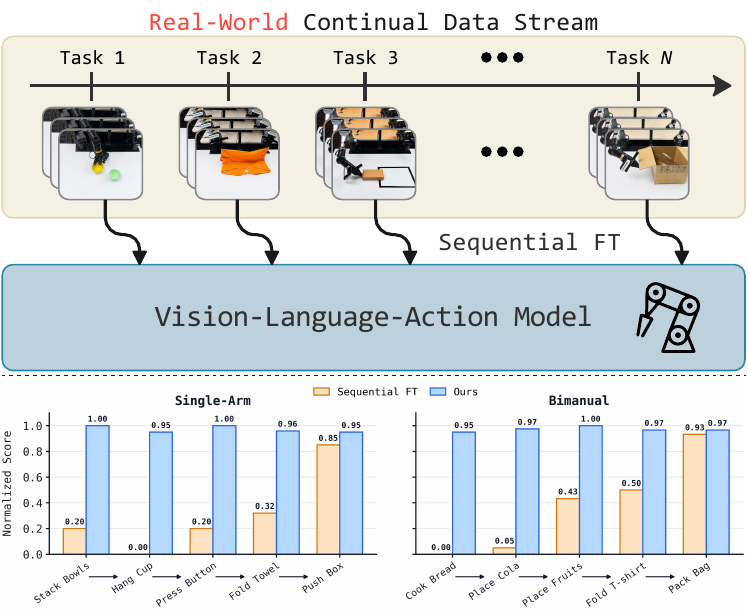}
    \caption{This paper investigates continual adaptation of VLA models under real-world robotic data streams. Our ER-based approach achieves robust continual learning performance across diverse manipulation tasks.}
    \label{fig:placeholder}
\end{figure}

Recent studies have begun to explore continual learning in VLA models from various perspectives. For example, \cite{liu2026pretrained} shows that large pretrained VLAs exhibit notable inherent resistance to forgetting, where model scale, pretraining priors, and even minimal replay buffers significantly enhance skill retention during sequential updates. Furthermore, \cite{hu2026simple} demonstrates that simple sequential fine-tuning combining pretrained VLAs, parameter-efficient adaptation, and on-policy reinforcement learning can overcome catastrophic forgetting while maintaining a favorable balance among retention, plasticity, and generalization. To address settings where historical data is constrained or unavailable, \cite{romer2026clare} proposes a dynamic adapter expansion framework that freezes existing model parameters and fine-tunes dedicated task-specific adapters to prevent parameter overwriting. Meanwhile, \cite{liu2026towards} explores reinforcement fine-tuning tailored for long-lived robots, using on-policy interactions to enable continuous adaptation of VLA models while avoiding the severe skill degradation common in standard supervised fine-tuning.

Despite these advances, existing continual learning evaluations for VLA models suffer from key limitations that systematically overestimate their robustness to forgetting. \textbf{First, simulation-based benchmarks lack the heterogeneity found in physical deployments.} Dominant benchmarks such as LIBERO \cite{liu2023libero} evaluate narrowly defined subtasks under fixed embodiments, environments, and action spaces. Such limited distribution shifts tend to underestimate forgetting, as retaining performance across highly correlated tasks does not guarantee generalization to heterogeneous real-world deployments. \textbf{Second, the observed resistance to forgetting may be confounded by overlap with pretraining data.} As noted by \cite{liu2026pretrained}, performance retention during sequential updates often stems from pretraining priors rather than algorithmic efficacy. Since standard benchmarks are frequently included in VLA pretraining corpora, high retention scores may simply reflect the reactivation of memorized skills rather than effective continual learning. \textbf{Third, standard evaluation protocols frequently violate causality.} These studies \cite{liu2026pretrained,hu2026simple} precompute global statistics (\eg, action normalization parameters) across the entire task stream before sequential training, thereby leaking future information into the learning process. Reported success is thus an artifact of an unrealistic, information-privileged regime unavailable in real-world deployment. These issues raise questions about whether current findings generalize to realistic settings and obscure whether forgetting in VLA models stems primarily from algorithmic failure or from overlooked factors, such as distributional inconsistency, normalization drift, and optimization imbalance under real-world sequential data.

Inspired by the discussions above, we shift the evaluation of continual learning in VLA models from simulation to physical environments under realistic deployment constraints. Our main contributions are threefold: 

\begin{itemize}
    \item We establish a real-world continual manipulation benchmark comprising ten sequential manipulation tasks (5 single-arm + 5 bimanual) characterized by high task diversity, including rigid-object pick-and-place, contact-rich pressing, and deformable-object folding, etc. To support standardized research, we have collected 4,000 high-quality demonstration trajectories and fully open-sourced the benchmark suite, dataset, and codebase\footnote{https://github.com/Agentic-Intelligence-Lab/ContinualVLA}.

    \item We systematically investigate the key factors governing real-world VLA adaptation, showing first that naive sequential supervised fine-tuning leads to severe catastrophic forgetting on real-world demonstration data streams. Then, we evaluate the experience replay-based approach, demonstrating that its performance significantly depends on proper configurations and highlighting the crucial role of action normalization strategies in maintaining stability during continual learning. 

    \item We successfully achieve robust continual learning across a 10-task heterogeneous real-world manipulation task stream. By synthesizing our empirical findings into a well-configured experience replay and action normalization pipeline, our approach effectively retains past capabilities while acquiring diverse skills, outperforming joint multi-task training under equivalent compute budgets in real-world deployment scenarios.
\end{itemize}

\section{Related Work}
\subsection{Vision-Language-Action Models}
VLA models extend vision-language backbones to physical manipulation by directly mapping multimodal inputs to executable action spaces. Early VLA architectures primarily employ autoregressive action tokenization, as exemplified by models such as OpenVLA~\cite{kim2024openvla}. More recent foundational policies, such as $\pi_0$~\cite{black2024pi_0} and $\pi_{0.5}$~\cite{black2025pi}, introduce continuous flow-matching heads to capture complex multimodal action distributions at high control frequencies. Alongside architectural scaling, recent research has explored diverse dimensions of policy learning, including diffusion-based manipulation policies~\cite{liu2025rdt}, cross-embodiment datasets~\cite{zheng2026xvla}, action-centric world modeling~\cite{worldvla}, and scalable post-training or reinforcement learning paradigms~\cite{llava-vla,sun2026atomvla}. Despite these advances in static multi-task pretraining and offline fine-tuning, current VLA frameworks are predominantly evaluated on fixed datasets instead of non-stationary task streams. In contrast to standard post-training protocols, our work focuses on the continual adaptation of VLAs under sequential, heterogeneous real-world data, systematically investigating the factors governing physical skill retention and policy stability.

\begin{figure*}[t!]
\centering
\includegraphics[width=\linewidth]{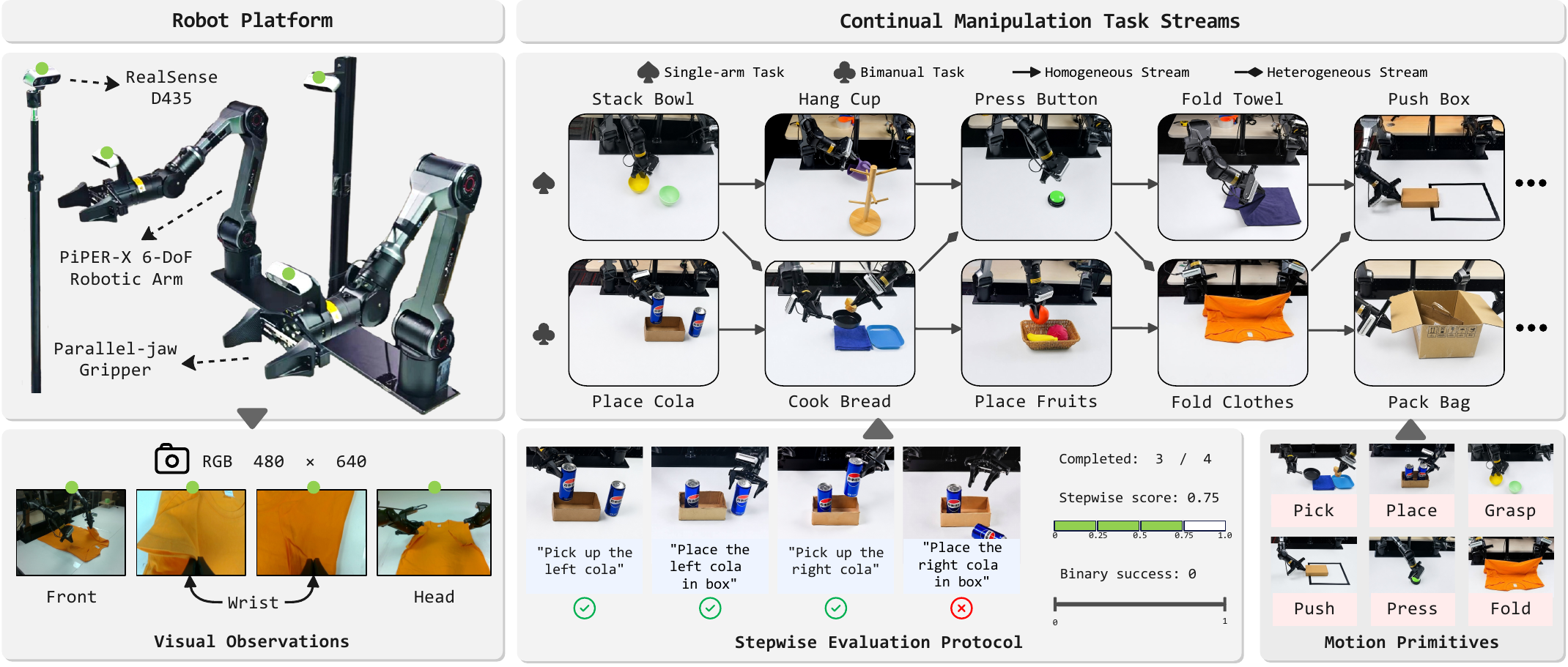}
\caption{
Overview of the robot platform, individual manipulation tasks, stepwise evaluation protocol, and two distinct continual task streams. These tasks span rigid-body manipulation, bimanual coordination, and complex deformable-object manipulation, requiring diverse motion primitives and providing a comprehensive benchmark for evaluating the continual learning capability of VLA models in real-world scenarios.} 
\label{fig:robot_setup}
\end{figure*}

\subsection{Continual Learning}
 Continual learning studies how agents acquire new capabilities sequentially while mitigating catastrophic forgetting under non-stationary data distributions \cite{de2021continual,wang2024comprehensive,shi2025continual}. Predominant paradigms of continual learning approaches include regularization-based methods \cite{serra2018overcoming,schwarz2018progress,ritter2018online}, architecture-based methods \cite{rusu2016progressive,mallya2018packnet,fernando2017pathnet}, optimization-based methods \cite{lopez2017gradient,chaudhry2018efficient,tang2021layerwise}, and replay-based methods \cite{rebuffi2017icarl,rolnick2019experience,buzzega2020dark}. Among these, experience replay is a widely used strategy that eliminates forgetting by storing and interleaving historical samples during sequential optimization. While pioneering works~\cite{liu2026pretrained,hu2026simple} evaluate VLA continual learning on simulated benchmarks, we focus on real-world adaptation using physical robot demonstration streams. In contrast to simulation, real-world sequential fine-tuning induces severe catastrophic forgetting. We show that policy stability under physical deployment constraints relies heavily on execution-level choices, specifically proper ER configurations and consistent action normalization strategies.

\section{Preliminaries}
\subsection{Continual Robot Learning}
We model the robot learning problem as a finite-horizon Markov decision process \cite{puterman1990markov}, defined by the tuple $\mathcal{M}=(\mathcal{S}, \mathcal{A}, \mathcal{T}, H, \mu_0, R)$. Here, $\mathcal{S}$ is the state space, $\mathcal{A}$ is the action space, $\mathcal{T}:\mathcal{S}\times\mathcal{A}\rightarrow\mathcal{S}$ is the transition function, $H$ is the finite episode horizon, $\mu_0$ is the initial state distribution, and $R:\mathcal{S}\times\mathcal{A}\rightarrow\mathbb{R}$ is the reward function. To evaluate the task success, we assume a sparse-reward setting and replace $R$ with a binary goal indicator $g:\mathcal{S}\rightarrow\{0,1\}$. Finally, the goal of the robot is to learn a policy $\pi$ that maximizes the expected return $J_{\rm RL}^\pi=\mathbb{E}_{(\bm{s}_t,\bm{a}_t)\sim\pi,\mu_0}\left[\sum_{t=1}^{H}g(\bm{s}_t)\right]$.

In the continual robot learning scenario, the robot learns a sequence of tasks $\{T^{k}\}_{k=1}^{K}$ with a single policy $\pi(\cdot|\bm{s};T)$ that conditions on the task. Notably, each task $T^k$ is defined by a unique initial state distribution $\mu_0^k$ and a goal predicate $g^k$, while all the tasks share the same $(\mathcal{S},\mathcal{A},\mathcal{T},H)$. Upon receiving the $k$-th task, the robot aims to maximize 
\begin{equation}
    J_{\rm CRL}^\pi=\frac{1}{k}\sum_{p=1}^{k}\left[\mathbb{E}_{(\bm{s}_t^p,\bm{a}_t^p)\sim \pi(\cdot;T^p),\mu^p_0}\left[ \sum_{t=1}^{H}g^{p}(\bm{s}_t^{p})\right]\right].
\end{equation}

\subsection{Continual Imitation Learning}
In this paper, we focus on the continual robot learning via imitation learning, especially the supervised fine-tuning (FT) of VLA models. Given expert demonstrations $\mathcal{D}^k=\{\tau_i^k\}_{i=1}^N$ for each task $T^{k}$, where each trajectory $\tau=\{(\bm{o}_{0},\bm{a}_0),\dots,(\bm{o}_{l^k},\bm{a}_{l^k})\},l^k\leq H$ consists of robot sensory inputs $\bm{o}_t$ and actions $\bm{a}_t$. Upon receiving the $k$-th task, our objective is to learn a policy that minimizes:
\begin{equation}
    J_{\rm BC}^\pi=\frac{1}{k}\sum_{p=1}^{k}\mathbb{E}_{(\bm{o}_t,\bm{a}_t)\sim\mathcal{D}^p}\left[\sum_{t=0}^{l^p}\mathcal{L}(\pi(\bm{o}_{\leq t};T^{p}),\bm{a}_{t}^{p})\right],
\end{equation}
where $\bm{s}_t\equiv\bm{o}_{\leq t}\triangleq(\bm{o}_0,\bm{o}_1,\dots,\bm{o}_t)$ is represented by the aggregated history of observations, $\mathcal{L}$ is the behavior cloning loss (\eg, the negative log-likelihood loss), and $\{\mathcal{D}^{p}:p<k\}$ is not fully available when learning $T^k$.


\begin{figure*}[ht!]
    \centering
    \includegraphics[width=\linewidth]{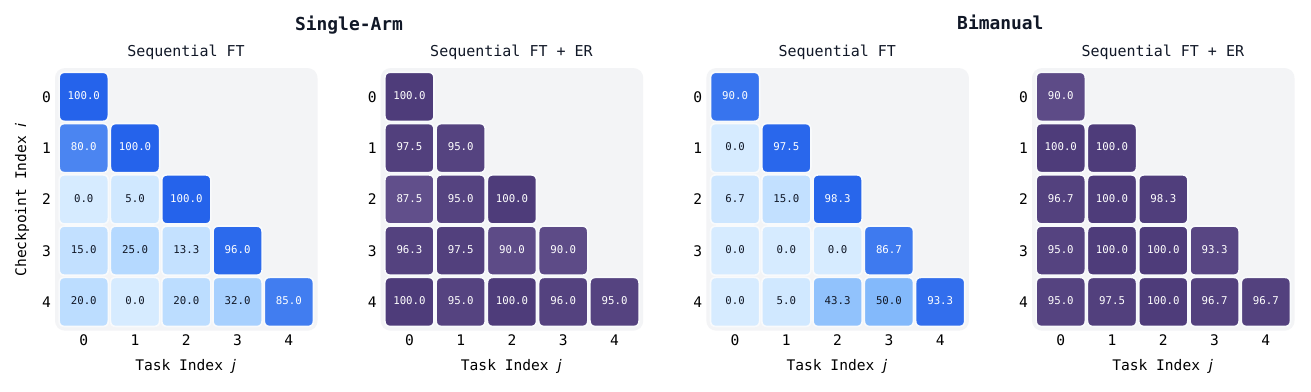}
    \caption{
Comparison of the continual learning performance between the naive sequential FT and ER on the two homogeneous task streams. While naive sequential FT suffers severe performance loss on real-world data streams, ER can effectively eliminate catastrophic forgetting in both single-arm and bimanual tasks.}
    \label{fig:four_continual_heatmaps}
\end{figure*}

\section{Real-World Continual Manipulation Benchmark}\label{sec:forgetting}

Unlike simulated benchmarks, which often feature idealized dynamics and limited environmental diversity, leading to an overestimation of model robustness, real-world robotic task streams present highly non-stationary data distributions, complex physical interactions, and unpredictable environmental disturbances. To evaluate how VLA models adapt under realistic deployment conditions, this section develops a real-world continual manipulation benchmark comprising 10 diverse manipulation tasks with substantial inter-task variation. 

\subsection{Robot Platform Setup} 
As shown in Figure~\ref{fig:robot_setup}, we conducted all experiments using two PiPER-X 6-DoF robotic arms, each equipped with a parallel-jaw gripper. Additionally, four RealSense D435 are deployed to capture the visual observations: one front camera, one head camera, and two wrist-mounted cameras. More detailed hardware configurations and real-world deployment setups are provided in the Appendix.

\subsection{Data Collection} 
A leader-follower pair of robotic arms is utilized in a teleoperation framework to acquire demonstration data. We collect 500 trajectories for each single-arm task, and 300 trajectories for each bimanual task. The trajectory includes robot action records and visual sensory inputs. Visual observations are captured at 30 Hz via four RGB cameras, each stream operating at a resolution of $480 \times 640$. Finally, a 35 GB dataset is derived and stored in the LeRobot format \cite{cadene2026lerobot}.

\subsection{Tasks Design}
As illustrated in Figure~\ref{fig:robot_setup}, we design 5 single-arm and 5 bimanual tasks that span rigid-body manipulation (\eg, stacking and hanging), bimanual coordination, and complex deformable-object manipulation (\eg, folding clothes). This task progression introduces substantial distributional shifts in kinematics, object geometry, and contact mechanics to rigorously test skill retention. Equipped with the designed tasks, we define two distinct continual task streams. First, the \textit{homogeneous task stream} evaluates sequential learning under homogeneous control modes by training the model on either the five single-arm tasks or the five bimanual tasks independently. Second, the \textit{heterogeneous task stream} alternates between single-arm and bimanual manipulations. This design allows us to isolate and study policy degradation under structurally similar dynamics versus the interference caused by abrupt transitions in robot kinematics and control modes. Finally, we provide complete task specifications and evaluation protocols in the Appendix.

\subsection{Stepwise Evaluation Protocol} 
Since real-world manipulation involves long-horizon execution and stochastic failures, binary success metrics are insufficient to distinguish different levels of task completion. A stepwise evaluation approach provides a more informative and reliable measure by recognizing intermediate progress, rather than focusing solely on the final outcome of success or failure. To evaluate the continual learning performance, we introduce three complementary metrics \cite{wang2024comprehensive}. (i) Denote by $c_{i,j}$ the score on the $i$-th task when evaluated using the model trained up to the $j$-th task. We report the \textbf{average score (AS)} across all tasks at the final checkpoint: $\mathrm{AS} = \frac{1}{K}\sum_{i=1}^{K} c_{i,K}.$ (ii) To quantify knowledge loss on previously learned tasks, we adopt the \textbf{backward transfer (BWT)} metric: $\mathrm{BWT}_k = \frac{1}{k-1}\sum_{j=1}^{k-1} \left(c_{j,k} - c_{j,j}\right),$ where a negative BWT indicates that performance degrades after learning subsequent tasks. The higher the value, the better. (iii) To assess whether sequential learning facilitates or hinders the acquisition of new tasks, the \textbf{forward transfer (FWT)} metric is introduced: $\mathrm{FWT}_k = \frac{1}{k-1}\sum_{j=2}^{k}\left(c_{j,j}-\bar{c}_j\right)$, where $\bar{c}_{j}$ is the single-task baseline performance. Positive $\mathrm{FWT}_i$ indicates that prior tasks facilitate learning new tasks, whereas negative $\mathrm{FWT}_i$ indicates plasticity loss.

\begin{table*}[]
\centering
\caption{Comparison of continual learning performance on real-world VLA task streams. Scores are evaluated after training on the final task and reported for all tasks.}
\label{tab:task_level_metrics}
\begin{tabular}{cccccccccc}
\toprule
\textbf{Type} & \textbf{Method} & $\mathcal{D}_1$ & $\mathcal{D}_2$ & $\mathcal{D}_3$ & $\mathcal{D}_4$ & $\mathcal{D}_5$ & \textbf{AS} ($\uparrow$) & \textbf{BWT} ($\uparrow$) & \textbf{FWT} ($\uparrow$) \\ \midrule
               & Baseline             & 100.0           & 97.5            & 100.0           & 52.0            & 85.0            & 86.9                          & --                         & --                      \\
Single-Arm    & Joint training  & 97.5            & 95.0            & 33.3            & 92.0            & 95.0            & 82.6                          & --                         & --                      \\
\textit{}     & Sequential FT   & 20.0            & 0.0             & 20.0            & 32.0            & 85.0            & 31.4                          &     -81.0                       &     +11.6                    \\
\textit{}     & ER              & 100.0           & 95.0            & 100.0           & 96.0            & 95.0            &  97.2                         &      +1.5                      &    +11.4                     \\ \midrule
\textit{}     & Baseline             & 90.0            & 90.0            & 96.7            & 76.7            & 86.7            & 88.0                          & --                         & --                      \\
Bimanual      & Joint training  & 80.0            & 82.5            & 93.3            & 73.3            & 86.7            &  83.2                         & --                         & --                      \\
\textit{}     & Sequential FT   & 0.0             & 5.0             & 43.3            & 50.0            & 93.3            &  38.3                         & -68.6                           &    +6.4                     \\
\textit{}     & ER              & 95.0            & 97.5            & 100.0           & 96.7            & 96.7            & 97.2                          &    +1.90                        &   +9.6                      \\ \bottomrule
\end{tabular}
\end{table*}

\section{Towards Robust Continual VLA Learning via Experience Replay}
Equipped with the developed benchmark, we first investigate whether the VLA models suffer from catastrophic forgetting when trained sequentially on real-world data streams. Furthermore, we investigate whether experience replay (ER) \cite{rolnick2019experience}, which interleaves historical demonstrations from a bounded buffer with the incoming data stream, can mitigate forgetting in continual VLA adaptation with real-world demonstration data. We systematically examine how key replay configurations, such as replay ratio and replay frequency, govern the overall behavior of the ER approach. In particular, we further demonstrate that selecting an appropriate action normalization strategy significantly affects the final performance of the ER approach.

\subsection{Continual Adaptation of VLA Models Under Real-World Data Streams}
\subsubsection{Model Baseline}
We adopt $\pi_{0.5}$ \cite{black2025pi} as our base VLA model and follow its standard supervised FT pipeline. To monitor performance degradation across the continual adaptation stages, we evaluate checkpoints by rolling out the policy on each task and computing normalized scores (0-100) based on task-specific rubrics. Complete training hyperparameters, optimization schedules, and baseline configurations are deferred to the Appendix.

\subsubsection{Experience Replay Setup}
To implement ER under practical data constraints, we maintain a replay buffer $\mathcal{B}$ with its capacity bounded by $N_\mathcal{B}=\rho_{\mathcal{B}}N_{\mathrm{ep}}$, where $\rho_{\mathcal{B}}\in(0,1]$ denotes the replay ratio and $N_{\mathrm{ep}}$ is the total number of collected episodes. To ensure balanced memory representation, this capacity is uniformly partitioned among the $K$ encountered tasks, allocating $\lfloor N_\mathcal{B}/K\rfloor$ episodes per task to prevent recent tasks from dominating the replay buffer during continual adaptation. During training, we regulate the stability-plasticity trade-off by dynamically constructing mini-batches from either the replay buffer $\mathcal{D}_{\mathrm{rep}}$ or current-task demonstrations $\mathcal{D}_{\mathrm{cur}}$. Specifically, these batches are selected according to the probability distribution $P(\mathcal{D}_{\mathrm{rep}})=f_r$ and $P(\mathcal{D}_{\mathrm{cur}})=1-f_r$, where $f_r\in[0,1]$ defines the replay frequency.

\subsubsection{Naive Sequential FT Causes Catastrophic Forgetting.}
As shown in Figure~\ref{fig:four_continual_heatmaps} and Table~\ref{tab:task_level_metrics}, naive sequential FT suffers from severe catastrophic forgetting across both single-arm and bimanual task streams. 
For the single-arm stream, upon completing sequential adaptation via task $\mathcal{D}_5$, the model's AS drops sharply from $86.9$ to $31.4$, resulting in a low BWT of $-81.0$. Early tasks experience near-total performance collapse, with AS on $\mathcal{D}_1$ and $\mathcal{D}_2$ falling to $20.0$ and $0.0$, respectively. A similar failure pattern emerges in the bimanual stream, where sequential FT achieves an AS of merely $38.3$ and a BWT of $-68.6$, losing the vast majority of previously learned capabilities. These results demonstrate that standard sequential FT fails under deployment-realistic streams. Crucially, due to the pronounced non-stationarity of real-world physical data, VLA models exhibit a higher degree of forgetting than previously reported in simulated benchmarks \cite{liu2026pretrained}.

\subsection{Well-Configured ER Effectively Eliminates Catastrophic Forgetting}

\subsubsection{Small Replay Budgets Preserve Previous Skills and Enable Transfer} 

ER effectively restores stability in continual learning, even under a minimal replay budget. As shown in Table~\ref{tab:task_level_metrics}, equipping sequential FT with ER achieves an AS of $97.2$ across both single-arm and bimanual streams, substantially drastically reducing forgetting to minimal degradation (BWT of $+1.5$ and $+1.9$, respectively). Remarkably, well-configured ER consistently outperforms joint multi-task training under equivalent computational budgets, which achieves an AS of only $82.6$ on the single-arm stream and $83.2$ on the bimanual stream. This superiority stems from the fact that joint training on heterogeneous physical tasks suffers from optimization conflicts and gradient interference when all skills are optimized simultaneously. In contrast, sequential learning with ER enables staged representation learning while maintaining historical stability through replay.

\begin{figure}[h!]
    \centering
    \includegraphics[width=\linewidth]{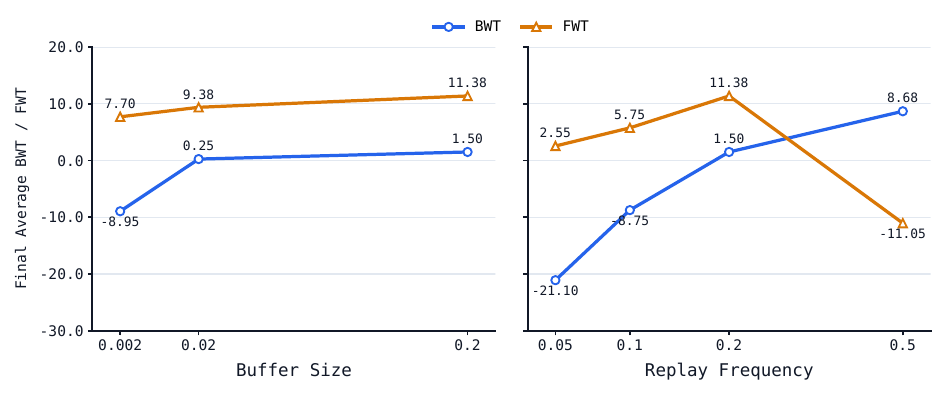}
    \caption{Sensitivity analysis of ER configurations. Increasing buffer capacity enhances retention, while replay frequency presents a stability–plasticity trade-off between insufficient (forgetting) and excessive (over-regularization) replay.}
    \label{fig:er_sensitivity}
\end{figure}

\subsubsection{Replay Frequency Balances Stability and Plasticity}
We evaluate ER sensitivity across buffer ratios $\rho_{\mathcal{B}} \in \{0.002, 0.02, 0.2\}$ and replay frequencies $f_r \in \{0.05, 0.1, 0.2, 0.5\}$. As shown in Figure~\ref{fig:er_sensitivity}, increasing buffer capacity improves retention, increasing final average BWT from $-8.95$ ($\rho_{\mathcal{B}} = 0.002$) to $+0.25$ ($\rho_{\mathcal{B}} = 0.02$) and $+1.50$ ($\rho_{\mathcal{B}} = 0.2$), proving that a modest buffer ($\rho_{\mathcal{B}} = 0.02$) is already highly effective. Conversely, replay frequency $f_r$ governs a distinct stability--plasticity trade-off. Low frequency ($f_r = 0.05$) leads to high forgetting ($\text{BWT} = -21.1$), whereas increasing $f_r$ to $0.1$ and $0.2$ progressively raise BWT to $-8.75$ and $+1.50$, respectively. Although aggressive replay ($f_r = 0.5$) achieves the highest BWT ($+8.68$), it over-regularizes the policy and impairs adaptation to new tasks. Consequently, $f_r = 0.2$ provides the optimal balance between historical skill retention and new-task plasticity.

\begin{figure}[t!]
    \centering
    \includegraphics[width=\linewidth]{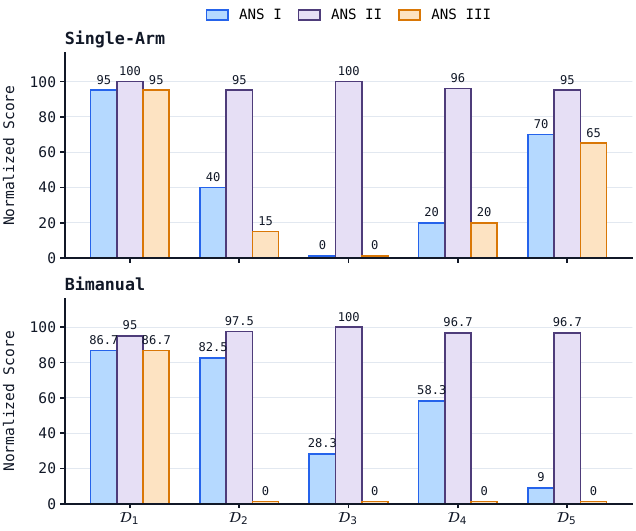}
    \caption{Comparison of the continual learning performance across three action normalization strategies. Here, \textbf{ANS \Rmnum{2}} achieves the best performance in both single-arm and bimanual task streams, highlighting the importance of maintaining action normalization consistency in the continual learning process.}
    \label{fig:ans}
\end{figure}

\subsection{Action Normalization Consistency Is Critical for Continual Adaptation}

Continual VLA learning requires a consistent action coordinate system throughout continual adaptation. Since future task data is unavailable, normalization statistics must be determined before training. In this paper, we design and test three distinct action normalization strategies (ANS). Specifically, \textbf{ANS \Rmnum{1}}: The policy is trained and tested on each task using its local, task-specific action statistics. \textbf{ANS \Rmnum{2}}: The normalization statistics of the first task are frozen and applied to the entire task stream for both training and testing. \textbf{ANS \Rmnum{3}}: The model is trained sequentially using each task's individual normalization statistics, yet the normalization statistics of the first task are utilized for all the testing.

We find that a fixed normalization space can support effective continual learning when it provides sufficient action coverage. Using the first-task statistics (\textit{stack bowls} for single-arm and \textit{cook bread} for bimanual sequences), models achieve stable optimization and strong performance across subsequent tasks. In contrast, allowing normalization statistics to vary across tasks disrupts cross-task alignment, preventing reliable knowledge transfer even when task-specific behaviors are successfully learned. These results indicate that maintaining a shared action space is more important than adapting normalization independently for each task.

\section{Continual Adaptation on Heterogeneous Task Stream}

Building upon the insights gained from evaluations of homogeneous task streams, which established effective ER configurations and the critical role of action normalization strategy, we scale our investigation to a significantly more challenging deployment scenario. In this section, we evaluate continual adaptation on a long-horizon, heterogeneous task stream comprising all ten manipulation tasks, executed in alternating order. Unlike homogeneous streams that preserve uniform robot kinematics, this heterogeneous protocol constantly interleaves single-arm and bimanual tasks. This setup introduces severe non-stationarity by forcing a single policy to adapt to a sequence characterized by abrupt, recurring transitions in control dimensionality, kinematic configurations, and physical contact dynamics. By testing the model on this complex task stream, we investigate whether our ER framework and action normalization strategies can mitigate cross-task interference and prevent policy collapse in the face of diverse physical-manipulation demands.

\begin{figure}[ht!]
    \centering
    \includegraphics[width=\linewidth]{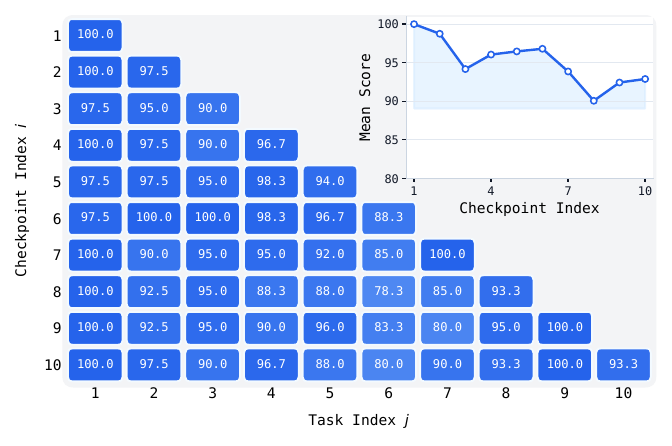}
    \caption{The continual learning performance on the heterogeneous task stream. Our method achieves stable continual learning across the long-horizon task sequence, with consistent retention and adaptation throughout all checkpoints.}
    \label{fig:heterogeneous_matrix}
\end{figure}

\begin{figure}[ht!]
    \centering
    \includegraphics[width=\linewidth]{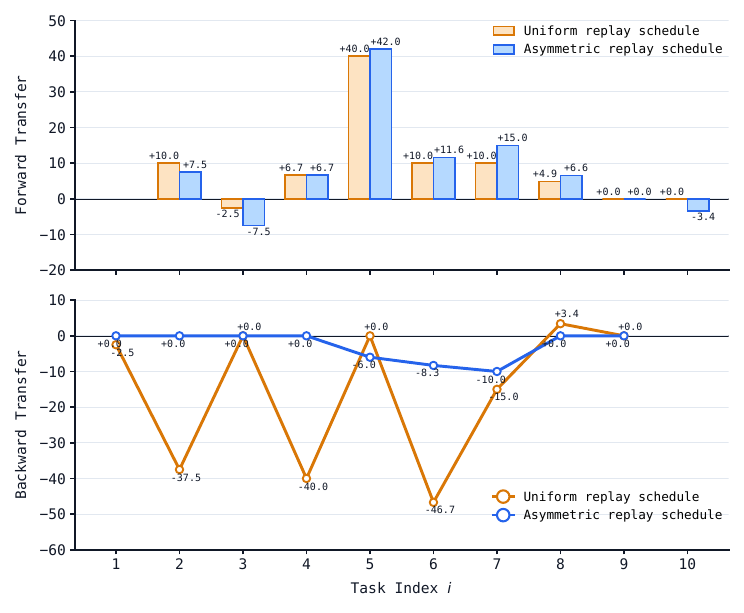}
    \caption{Trace of the BWT and FWT metrics across the heterogeneous task stream. A uniform replay frequency across all tasks degrades performance in bimanual tasks, whereas an asymmetric replay schedule effectively eliminates forgetting.}
    \label{fig:hetero_nbt_ft}
\end{figure}

\begin{figure*}[t!]
    \centering
    \includegraphics[width=\linewidth]{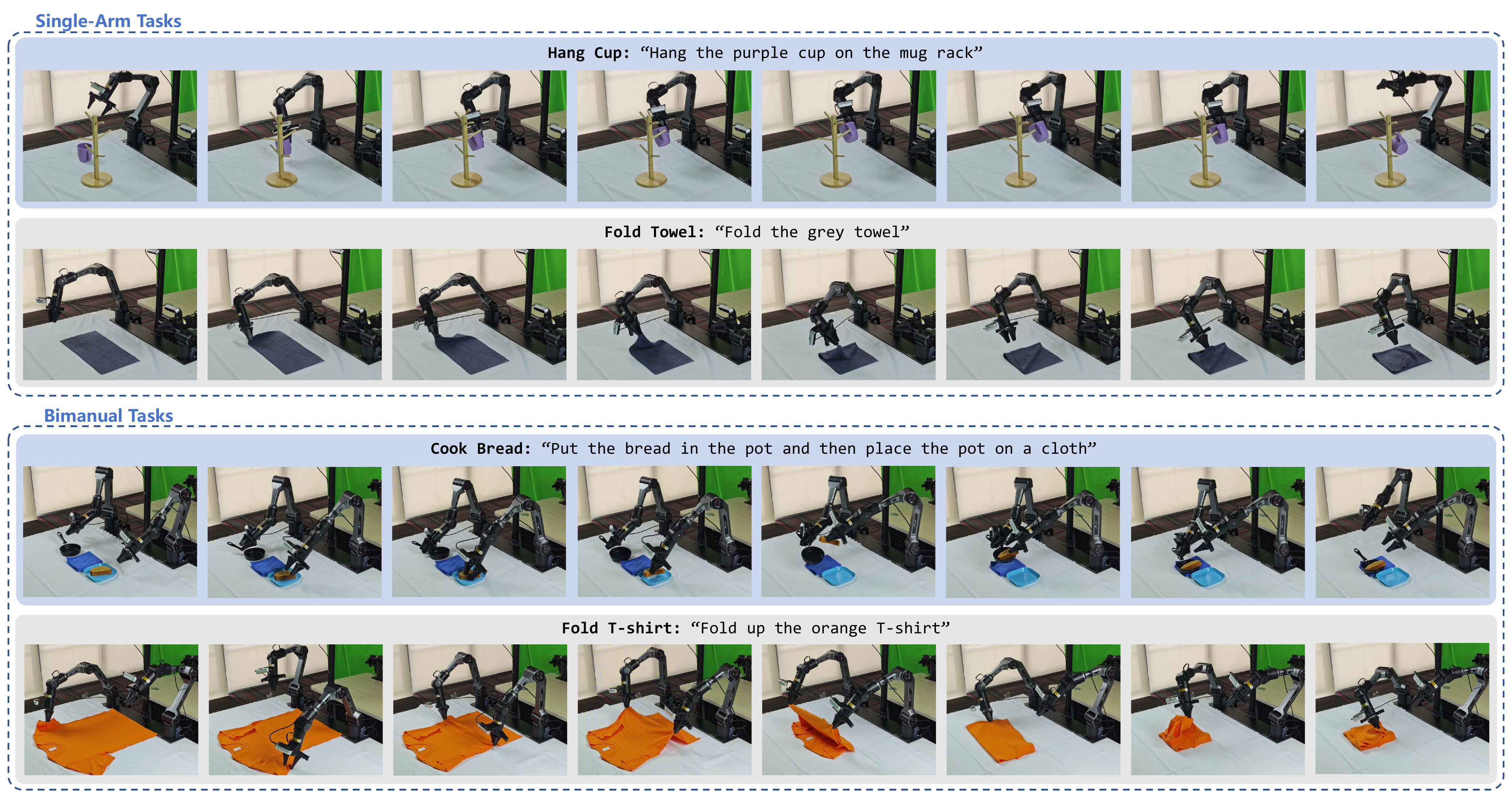}
    \caption{Qualitative rollout snapshots evaluated after completing the full 10-task heterogeneous continual learning stream. Representative keyframes showcasing physical robot execution on selected single-arm and bimanual tasks. The final adapted policy demonstrates robust execution and skill retention across diverse manipulation primitives after sequential training across all 10 tasks, without suffering from catastrophic forgetting or action collapse.}
    \label{fig:hetero_demos}
\end{figure*}

\paragraph{Overall Continual Performance} We first evaluate the overall efficacy of the adapted policy across the entire 10-task heterogeneous stream. As illustrated in the performance matrix (Figure~\ref{fig:heterogeneous_matrix}) and the rollout snapshots (Figure~\ref{fig:hetero_demos}), the policy successfully acquires new manipulation skills sequentially while retaining previously learned behaviors. Rather than suffering from catastrophic forgetting or action mode collapse, the agent maintains robust execution across both single-arm and bimanual tasks throughout the adaptation process. To quantitatively analyze the adaptation dynamics, Figure~\ref{fig:hetero_nbt_ft} presents the task-by-task BWT and FWT. The bar components show the FWT for each task, maintaining a remarkably high average of $+8.7$ by the final stage. Simultaneously, the line plot traces the BWT across the task sequence, yielding an average BWT of $-2.7$. This dual-metric analysis reveals that our ER-based strategy effectively balances stability and plasticity, minimizing backward interference while facilitating positive forward transfer even across abrupt kinematic transitions.

\paragraph{Asymmetric Replay Schedule for Kinematic Transitions}
A key empirical insight from our heterogeneous stream evaluation is that applying a uniform replay frequency across all tasks yields suboptimal stability. Specifically, when employing the default homogeneous configuration ($f_r = 0.2$), the policy experiences severe performance degradation on previously learned bimanual tasks when adapting to subsequent single-arm tasks. This heightened vulnerability arises because bimanual manipulation inherently requires higher-dimensional spatial coordination and exhibits complex control dynamics, making its task representations particularly fragile to parameter overwriting. Furthermore, when the model transitions to single-arm tasks, optimization focuses exclusively on a single arm, generating gradient updates that actively suppress and unlearn the action outputs associated with the inactive arm. To counter this cross-kinematic interference, we introduce an asymmetric replay schedule. Specifically, we increase the overall replay frequency to $f_r=0.3$ and bias the replay buffer sampling toward bimanual experiences at a $2:1$ ratio relative to single-arm experiences. This design ensures that the model anchors the complex dual-arm coordination patterns with sufficient frequency, while still retaining adequate plasticity to adapt to subsequent single-arm tasks.

\section{Conclusion}
In this work, we establish a real-world continual manipulation benchmark to systematically evaluate VLA models' adaptation under physical deployment constraints. Through extensive experiments on heterogeneous manipulation tasks, we demonstrate that naive sequential FT suffers from severe catastrophic forgetting due to the high non-stationarity of real-world physical data. We show that well-configured ER effectively eliminates skill degradation while enabling positive forward transfer. Surprisingly, our results reveal that continual learning-based training can outperform joint multi-task training under equivalent computational budgets. 

Despite these insights, our study presents several limitations that open promising avenues for future research. First, our investigation focuses primarily on replay-based adaptation within a fixed physical platform setup. Extending continual adaptation mechanisms to cross-embodiment setups, broader task distributions, and open-ended long-horizon scenarios remains an unaddressed challenge. Second, while ER effectively maintains policy stability, storing historical demonstration data may increase memory demands over prolonged, lifelong deployment. Future work could explore parameter-efficient modular adapters, exemplar-free architectures, and active memory management. This paper is expected to inspire more subsequent research toward building scalable, robust, and truly lifelong embodied agents.

\section*{Acknowledgements}
This work is supported by the HK UGC Early Career Scheme under Grant No. HKU 27205826. We also thank INFIFORCE for providing the computing resources.

\appendix

\section{Real-World Continual Manipulation Benchmark}\label{appendix:benchmark}




\subsection{Single-Arm Task Details}

\paragraph{Stack Bowls ($\mathcal{D}_1$)}
The robot must locate a yellow bowl, grasp it, and stack it stably on a green bowl. The task is evaluated using four intermediate checkpoints, with a maximum score of 4:

\begin{enumerate}[leftmargin=*,itemsep=1pt,topsep=3pt]
    \item The gripper approaches the yellow bowl within \(3\,\mathrm{cm}\).
    \item The robot successfully grasps and lifts the yellow bowl.
    \item The robot moves the yellow bowl above the green bowl.
    \item The robot releases the yellow bowl such that it remains stably stacked on the green bowl.
\end{enumerate}
A penalty of \(-0.5\) is applied if the green target bowl is knocked over during execution.

\paragraph{Hang Cup ($\mathcal{D}_2$)}
The robot must grasp a purple cup and hang it on a mug rack. The task is evaluated using four intermediate checkpoints, with a maximum score of 4:

\begin{enumerate}[leftmargin=*,itemsep=1pt,topsep=3pt]
    \item The robot correctly locates and grasps the cup.
    \item The robot moves the cup within \(3\,\mathrm{cm}\) of the target peg on the mug rack.
    \item The robot performs a valid hanging attempt using an inward insertion motion.
    \item The cup is successfully hung by its handle and remains on the rack after release.
\end{enumerate}

\paragraph{Press Button ($\mathcal{D}_3$)}
The robot must identify a green button, move the end-effector toward it, and press it successfully. The task is evaluated using three intermediate checkpoints, with a maximum score of 3:

\begin{enumerate}[leftmargin=*,itemsep=1pt,topsep=3pt]
    \item The robot correctly localizes the button and aligns the end-effector with an appropriate pressing orientation.
    \item The end-effector moves within \(3\,\mathrm{cm}\) of the button.
    \item The robot successfully presses and activates the button.
\end{enumerate}

\paragraph{Fold Towel ($\mathcal{D}_4$)}
The robot must fold a grey towel in half and tidy its corners and edges. The task is evaluated using five intermediate checkpoints, with a maximum score of 5:

\begin{enumerate}[leftmargin=*,itemsep=1pt,topsep=3pt]
    \item The robot correctly locates the top-right corner of the towel.
    \item The robot successfully grasps the top-right corner.
    \item The robot moves the grasped corner toward the corresponding top-left corner.
    \item The robot correctly locates and manipulates the bottom-right corner.
    \item The robot completes the fold and performs the final edge-alignment and tidying behavior.
\end{enumerate}
The final checkpoint is awarded only when the towel remains folded and the corresponding corners and edges are reasonably aligned.

\paragraph{Push Box ($\mathcal{D}_5$)}
The robot must establish contact with a box and push it into a designated target area. The gripper remains closed throughout the pushing process. The task is evaluated using four intermediate checkpoints, with a maximum score of 4:

\begin{enumerate}[leftmargin=*,itemsep=1pt,topsep=3pt]
    \item The robot correctly localizes the box and approaches an appropriate pushing surface.
    \item The closed gripper establishes effective contact with the box.
    \item The robot pushes the box toward the target area while maintaining a valid pushing motion.
    \item The box reaches and remains within the designated target area.
\end{enumerate}

\subsection{Bimanual Task Details}

\paragraph{Cook Bread ($\mathcal{D}_1$)}
The robot must use the left arm to manipulate a piece of bread and the right arm to manipulate a pot. It must place the bread into the pot and then place the pot on the designated towel. The task is evaluated using six intermediate checkpoints, with a maximum score of 6:

\begin{enumerate}[leftmargin=*,itemsep=1pt,topsep=3pt]
    \item The left arm successfully grasps and lifts the bread.
    \item The right arm successfully grasps and lifts the pot.
    \item The left arm moves the bread above the center of the designated towel.
    \item The right arm moves the pot beneath the bread and above the center of the towel.
    \item The left arm places the bread inside the pot.
    \item The right arm places the pot containing the bread on the designated towel.
\end{enumerate}
A penalty of \(-0.5\) is applied if the center of the pot is more than \(3\,\mathrm{cm}\) from the center of the towel after placement.

\paragraph{Place Cola ($\mathcal{D}_2$)}
The robot must sequentially use the right and left arms to place two cola cans into a paper box. The task is evaluated using four intermediate checkpoints, with a maximum score of 4:

\begin{enumerate}[leftmargin=*,itemsep=1pt,topsep=3pt]
    \item The right arm successfully grasps the cola can on the right.
    \item The right arm places the right cola can inside the paper box.
    \item The left arm successfully grasps the cola can on the left.
    \item The left arm places the left cola can inside the paper box.
\end{enumerate}
A penalty of \(-0.5\) is applied for each cola can knocked over during execution. An additional penalty of \(-0.5\) is applied if the robot severely compresses or deforms the paper box while placing a can.

\paragraph{Place Fruits ($\mathcal{D}_3$)}
The robot must place a dragon fruit, a banana, and a persimmon into a basket in the specified order. The task is evaluated using six intermediate checkpoints, with a maximum score of 6:

\begin{enumerate}[leftmargin=*,itemsep=1pt,topsep=3pt]
    \item The left arm successfully grasps the dragon fruit.
    \item The left arm places the dragon fruit inside the basket.
    \item The right arm successfully grasps the banana.
    \item The right arm places the banana inside the basket.
    \item The right arm successfully grasps the persimmon.
    \item The right arm places the persimmon inside the basket.
\end{enumerate}
A penalty of \(-0.5\) is applied if the right arm begins grasping another fruit before the left arm has placed the dragon fruit into the basket. A penalty of \(-1\) is applied if the right arm grasps the persimmon before the banana.

\paragraph{Fold T-shirt ($\mathcal{D}_4$)}
The robot must coordinate both arms to fold an orange T-shirt into one-quarter of its original size. The task is evaluated using five intermediate checkpoints, with a maximum score of 5:

\begin{enumerate}[leftmargin=*,itemsep=1pt,topsep=3pt]
    \item The robot successfully folds the right sleeve inward.
    \item The robot successfully folds the left sleeve inward.
    \item The two arms synchronously grasp the T-shirt near the two sides of its center.
    \item The robot folds the T-shirt in half.
    \item The robot folds the T-shirt once more into one-quarter of its original size.
\end{enumerate}
A penalty of \(-0.5\) is applied if a sleeve is only partially folded. An additional penalty of \(-0.5\) is applied if the T-shirt becomes visibly disordered after the first body fold.

\paragraph{Pack Bag ($\mathcal{D}_5$)}
The robot must use the right arm to place a paper bag into a cardboard box and then coordinate both arms to close the two box lids. The task is evaluated using six intermediate checkpoints, with a maximum score of 6:

\begin{enumerate}[leftmargin=*,itemsep=1pt,topsep=3pt]
    \item The right arm successfully grasps and lifts the paper bag.
    \item The right arm places the paper bag inside the cardboard box.
    \item The right arm moves beneath the right lid and assumes a valid pose for lifting and closing it.
    \item The left arm moves beneath the left lid and assumes a valid pose for lifting and closing it.
    \item The right arm closes the right lid.
    \item The left arm closes the left lid.
\end{enumerate}
Full task success requires the paper bag to remain inside the cardboard box and both lids to be closed.

\subsection{Task Stream Configurations}

The task order in each stream is as follows.

\paragraph{Single-Arm Task Stream}

\begin{tasklist}
    \item \textbf{Stack Bowls:} Stack the yellow bowl on the green bowl.
    \item \textbf{Hang Cup:} Hang the purple cup on the mug rack.
    \item \textbf{Press Button:} Press the green button.
    \item \textbf{Fold Towel:} Fold the grey towel.
    \item \textbf{Push Box:} Push the box into the designated area.
\end{tasklist}

\paragraph{Bimanual Task Stream}

\begin{tasklist}
    \item \textbf{Cook Bread:} Put the bread in the pot and then place the pot on a cloth.
    \item \textbf{Place Cola:} Place the cola into the paper box.
    \item \textbf{Place Fruits:} Pick up the three fruits and place them into the basket one by one.
    \item \textbf{Fold T-shirt:} Fold up the orange T-shirt.
    \item \textbf{Pack Bag:} Pick up the paper bag, put it into the cardboard box, and close the lid.
\end{tasklist}

\paragraph{Heterogeneous Task Stream}

\begin{tasklist}
    \item \textbf{Stack Bowls} (Single-Arm)
    \item \textbf{Place Cola} (Bimanual)
    \item \textbf{Hang Cup} (Single- Arm)
    \item \textbf{Cook Bread} (Bimanual)
    \item \textbf{Fold Towel} (Single-Arm)
    \item \textbf{Fold T-shirt} (Bimanual)
    \item \textbf{Push Box} (Single-Arm)
    \item \textbf{Pack Bag} (Bimanual)
    \item \textbf{Press Button} (Single-Arm)
    \item \textbf{Place Fruits} (Bimanual)
\end{tasklist}

\subsection{Stepwise Evaluation Protocol}

Binary success provides limited information when a policy completes only part of a manipulation task, particularly for long-horizon and bimanual behaviors. We therefore adopt a task-specific stepwise evaluation protocol that assigns partial credit for completing intermediate physical subgoals.

For a task containing \(M\) evaluation checkpoints, each completed checkpoint contributes one point, resulting in a maximum performance score of \(M\). The score is divided by \(M\) to obtain a normalized score ranging from 0 to 100. Any task-specific penalty is subtracted from the raw score before normalization. Obtaining all task scores requires completing all checkpoints and meeting the final placement, activation, alignment, or stability condition. A checkpoint is awarded only when its corresponding physical subgoal is visibly completed. Merely attempting a motion without producing the intended physical outcome does not receive credit.

\section{Details of Model Training}\label{appendix:model-training}

\subsection{Model Baseline}

Our experiments are based on $\pi_{0.5}$~\cite{black2025pi}, a flow-matching Vision-Language-Action (VLA) model developed by Physical Intelligence. The model comprises three main components: a SigLIP So400m vision encoder (${\sim}$400M parameters), a PaliGemma 2B language backbone (${\sim}$2B parameters) for cross-modal fusion, and a Gemma 300M Action Expert (${\sim}$311M parameters, 18-layer transformer with adaRMSNorm) that predicts action chunks via conditional flow matching. The total model size is approximately 2.7 billion parameters. All parameters are trainable; no components are frozen during continual learning. For full architectural details, we refer the reader to the original $\pi_{0.5}$ publication.

\subsection{Experimental Configuration}
\paragraph{FT} is standard single-task supervised fine-tuning: the $\pi_{0.5}$ base model is trained on one task for 4,000 steps. This serves as the reference point for individual task performance, free from interference from other tasks.

\paragraph{Joint training} mixes all tasks in a single training run with equal task sampling. The default behavior of $\pi_{0.5}$'s multi-dataset training is to concatenate all data and sample proportionally, which causes larger datasets to receive more training steps. To ensure a fair comparison, we apply inverse data-size weighting so that each task contributes equally to every batch regardless of its dataset size. The model is trained for 24,000 total steps, which aligns with the total step budget of ER (see below).

\paragraph{Sequential FT} trains tasks sequentially without any replay mechanism. Each task is trained for 4,000 steps, with the model initialized from the previous task's checkpoint. This represents the standard sequential fine-tuning baseline, in which catastrophic forgetting is expected.

\paragraph{Sequential FT+ER} adds experience replay to the sequential training protocol. After each task, a replay buffer is built, and prior-task buffers are downsampled as described in Section~\ref{app:replay-management}. The total step budget is $4{,}000 + 5{,}000 \times 4 = 24{,}000$, matching joint training exactly. This ensures that any performance difference between joint training and ER is attributable to the training paradigm (mixed versus sequential with replay) rather than the total number of gradient updates.

Finally, all experiments share the following experimental configuration:

\begin{table}[H]
\centering
\caption{Shared experimental configurations.}
\label{tab:config}
\small
\begin{tabular}{@{}p{0.32\columnwidth}p{0.62\columnwidth}@{}}
\toprule
\textbf{Hyperparameter} & \textbf{Value} \\
\midrule

\# GPUs 
& $8\times$ NVIDIA H20 (96~GB VRAM) \\

Batch size 
& 128 (global) / 16 (per GPU) \\

Optimizer 
& AdamW ($\beta_1=0.9$, $\beta_2=0.95$) \\

Precision
& bfloat16 \\

Learning rate 
& $5\times10^{-5}$ (peak) \\

LR schedule & Cosine decay with linear warmup \\

Warmup steps 
& 200 \\

Training steps 
& 4{,}000 \\

Gradient clipping 
& $\leq 1.0$ (global norm) \\

EMA decay 
& 0.998 \\

Base model 
& $\pi_{0.5}$ checkpoint \\

Random seed 
& 42 \\

Action normalization

& Quantile-based scaling ($q_{0.01}$--$q_{0.99}$) \\

Action chunk size
& 10 \\

\# Input images
& 4 (1 head camera + 1 third-person front camera + 2 wrist cameras) \\

Input image size
& 224 $\times$ 224 px \\

\# Patch tokens
& 256 (16 $\times$ 16) \\

Patch token size
& 14 $\times$ 14 px \\

\bottomrule
\end{tabular}
\end{table}


\subsection{Experience Replay Configurations}

\subsubsection{Replay Buffer Setup}
\label{app:replay-management}

We employ episode-level experience replay to mitigate catastrophic forgetting. The replay buffer $\mathcal{B}$ stores entire episodes rather than individual transitions, preserving intra-trajectory temporal coherence. Let $N_{\mathrm{ep}}^{(k)}$ denote the number of episodes in task $k$.

\paragraph{Buffer budget} The replay buffer ratio $\rho_{\mathcal{B}} = 0.2$ controls the total fraction of data retained across all previously learned tasks. The budget is shared equally: after $N$ tasks, each task contributes $\rho_{\mathcal{B}} / N$ of its episodes to $\mathcal{B}$.

\paragraph{Buffer construction} After training on task $k$ completes, a buffer is built by randomly selecting $K = \max(1, \lfloor N_{\mathrm{ep}}^{(k)} \times \rho_{\mathcal{B}} / k \rfloor)$ episodes. The selected episode indices and their contained step indices are recorded. A fixed seed ensures reproducibility.

\paragraph{Buffer downsampling} When a new buffer is added for task $k$, all existing buffers for tasks $1, \ldots, k-1$ are downsampled proportionally: for each prior task $i$, the target count becomes $K' = \max(1, \lfloor N_{\mathrm{ep}}^{(i)} \times \rho_{\mathcal{B}} / k \rfloor)$, and excess episodes are randomly discarded. Downsampling only touches the stored indices and never re-reads the original data, making it computationally negligible.

\begin{algorithm}[ht!]
\caption{Continual Adaptation with ER}
\label{alg:er}
\small
\begin{algorithmic}[1]
\REQUIRE Tasks $\{\mathcal{T}_1,\ldots,\mathcal{T}_N\}$ and base model $\theta_0$
\REQUIRE Steps $K$, buffer ratio $\rho_{\mathcal{B}}$, frequency $f_r$
\ENSURE Trained model $\theta$
\STATE $\theta \leftarrow \theta_0$ \COMMENT{$\pi_{0.5}$ base model}
\STATE $\mathcal{B} \leftarrow \emptyset$
\FOR{$k=1$ \TO $N$}
    \STATE $\mathcal{D}_k \leftarrow \textsc{LoadDataset}(\mathcal{T}_k)$
    \IF{$k = 1$ \OR $\mathcal{B} = \emptyset$}
        \STATE $S \leftarrow K$
    \ELSE
        \STATE $S \leftarrow \lfloor K/(1-f_r) \rfloor$
    \ENDIF
    \FOR{$t=0$ \TO $S-1$}
        \IF{$\mathcal{B}\neq\emptyset$ \AND $\textsc{Random}(t)<f_r$}
            \STATE $\text{batch}\leftarrow \textsc{Sample}(\mathcal{B})$
        \ELSE
            \STATE $\text{batch}\leftarrow \textsc{Sample}(\mathcal{D}_k)$
        \ENDIF
        \STATE $\theta\leftarrow
        \textsc{AdamW\_Step}(\theta,\text{batch})$
    \ENDFOR
    \FOR{$i=1$ \TO $k$}
        \STATE $n_i\leftarrow\max\!\left(1,\left\lfloor
        N_{\mathrm{ep}}^{(i)}\rho_{\mathcal{B}}/k\right\rfloor\right)$
    \ENDFOR
    \FOR{$i=1$ \TO $k-1$}
        \STATE $\mathcal{B}_i\leftarrow
        \textsc{Downsample}(\mathcal{B}_i,n_i)$
    \ENDFOR
    \STATE $\mathcal{B}_k\leftarrow
    \textsc{RandomEpisodes}(\mathcal{D}_k,n_k)$
    \STATE $\mathcal{B}\leftarrow
    \textsc{Concat}(\mathcal{B}_1,\ldots,\mathcal{B}_k)$
\ENDFOR
\RETURN $\theta$
\end{algorithmic}
\end{algorithm} 

The following table provides an evolution example for a 5-task sequence, where $N_{\mathrm{ep}}$ denotes the typical number of episodes per task. The total buffer size $N_{\mathcal{B}} = \rho_{\mathcal{B}} N_{\mathrm{ep}}$ remains constant throughout training.
\begin{table}[h!]
\centering
\caption{Evolution example of the replay buffer over a 5-task sequence.}
\label{tab:buffer}
\small
\begin{tabular}{@{}p{0.21\columnwidth}p{0.34\columnwidth}p{0.37\columnwidth}@{}}
\toprule
\textbf{Stage} & \textbf{Per-task fraction} & \textbf{Total size $N_{\mathcal{B}}$} \\
\midrule
After Task 1 & $\rho_{\mathcal{B}} / 1 = 20\%$ & $0.2 \, N_{\mathrm{ep}}$ \\
After Task 2 & $\rho_{\mathcal{B}} / 2 = 10\%$ & $0.2 \, N_{\mathrm{ep}}$ \\
After Task 3 & $\rho_{\mathcal{B}} / 3 \approx 6.7\%$ & $0.2 \, N_{\mathrm{ep}}$ \\
After Task 4 & $\rho_{\mathcal{B}} / 4 = 5\%$ & $0.2 \, N_{\mathrm{ep}}$ \\
After Task 5 & $\rho_{\mathcal{B}} / 5 = 4\%$ & $0.2 \, N_{\mathrm{ep}}$ \\
\bottomrule
\end{tabular}
\end{table}

\subsubsection{Replay Frequency Setup}

At each training step, the replay buffer $\mathcal{B}$ is sampled with probability $f_r = 0.2$:


\begin{equation*}
\text{Sample} =
\begin{cases}
\text{from } B, & \text{if } \text{Random}(t) < f_r, \\
\text{from the current task}, & \text{otherwise},
\end{cases}
\label{eq:sampling}
\end{equation*}
where the random number generator is seeded with the global step index $t$, ensuring identical decisions across all DDP ranks at each step.

\paragraph{Activation} Replay is disabled during task 1 (no prior tasks exist) and activates from task 2 onward. At task $k$, $\mathcal{B}$ is formed by concatenating buffer datasets from tasks $1, \ldots, k-1$, each mapping stored episode indices back to the original dataset. The replay data loader shares the same batch size and distributed sampling configuration as the current task's data loader.

\paragraph{Step scaling} To keep the number of new-task gradient updates constant with and without replay, the total steps per task are scaled by $1 / (1 - f_r)$. With $f_r = 0.2$, the total is $4{,}000 / 0.8 = 5{,}000$ steps, of which approximately 4,000 use current-task data and 1,000 use replay data.

Finally, we summarize the complete ER training loop in  Algorithm~\ref{alg:er}.

\bibliography{ref}
\bibliographystyle{ieeetr}

\end{document}